\documentclass[conference]{IEEEtran}
\usepackage{float}
\IEEEoverridecommandlockouts
\usepackage{cite}
\usepackage{float}
\usepackage[utf8]{inputenc}
\usepackage{url}
\usepackage{amsmath,amssymb,amsfonts}
\usepackage{algorithmic}
\usepackage{graphicx}
\usepackage{textcomp}
\usepackage{xcolor}
\usepackage{multirow}
\usepackage{amsmath}
\def\BibTeX{{\rm B\kern-.05em{\sc i\kern-.025em b}\kern-.08em
    T\kern-.1667em\lower.7ex\hbox{E}\kern-.125emX}}
\begin{document}

\title{{Integrating Travel Behavior Forecasting and Generative Modeling for Predicting Future Urban Mobility and Spatial Transformations\\
{\footnotesize \textsuperscript{}}
}
\thanks{This research was funded by the Economic Diversifying Research Fund (EDRF). Funding number-FAR0037938}
}

\author{\IEEEauthorblockN{1\textsuperscript{st} Eugene Denteh}
\IEEEauthorblockA{\textit{Dept. of Civil, Construction, and Environmental Engineering} \\
\textit{North Dakota State University}\\
Fargo, USA \\
eugene.denteh@ndsu.edu}
~\\
\and
\IEEEauthorblockN{1\textsuperscript{st} Andrews Danyo}
\IEEEauthorblockA{\textit{Dept. of Civil, Construction, and Environmental Engineering} \\
\textit{North Dakota State University}\\
Fargo, USA \\
andrews.danyo@ndsu.edu}
~\\
\and
\IEEEauthorblockN{3\textsuperscript{rd} Joshua Kofi Asamoah}
\IEEEauthorblockA{\textit{Dept. of Civil, Construction, and Environmental Engineering} \\
\textit{North Dakota State University}\\
Fargo, USA \\
joshua.asamoah@ndsu.edu}
~\\
\and
\IEEEauthorblockN{4\textsuperscript{th} Blessing Agyei Kyem}
\IEEEauthorblockA{\textit{Dept. of Civil, Construction, and Environmental Engineering} \\
\textit{North Dakota State University}\\
Fargo, USA \\
blessing.agyeikyem@ndsu.edu}

\and
\IEEEauthorblockN{5\textsuperscript{th} Twitchell Addai}
\IEEEauthorblockA{\textit{Dept. of Civil, Construction, and Environmental Engineering} \\
\textit{North Dakota State University}\\
Fargo, USA \\
twitchell.addai@ndsu.edu}

\and
\IEEEauthorblockN{6\textsuperscript{th} Armstrong Aboah*}
\IEEEauthorblockA{\textit{Dept. of Civil, Construction, and Environmental Engineering} \\
\textit{North Dakota State University}\\
Fargo, USA \\
armstrong.aboah@ndsu.edu}
*Corresponding author
}

\maketitle

\begin{abstract}
Transportation planning plays a critical role in shaping urban development, economic mobility, and infrastructure sustainability. However, traditional planning methods often struggle to accurately predict long-term urban growth and transportation demands. This may sometimes result in infrastructure demolition to make room for current transportation planning demands. This study integrates a Temporal Fusion Transformer to predict travel patterns from demographic data with a Generative Adversarial Network to predict future urban settings through satellite imagery. The framework achieved a 0.76 R² score in travel behavior prediction and generated high-fidelity satellite images with a Structural Similarity Index of 0.81. The results demonstrate that integrating predictive analytics and spatial visualization can significantly improve the decision-making process, fostering more sustainable and efficient urban development. This research highlights the importance of data-driven methodologies in modern transportation planning and presents a step toward optimizing infrastructure placement, capacity, and long-term viability.
\end{abstract}


\section{Introduction}
Transportation infrastructure is fundamental to social development, economic growth, social connectivity, and urban expansion \cite{Wan_2024}. As cities grow, strategic road developments and public transit systems shape urban landscapes and influence settlement patterns \cite{Kashef_2020}. These developments often catalyze the development of new commercial and residential districts. Effective transportation planning has historically facilitated the movement of people and goods, reduced congestion, and improved the quality of life in urban areas \cite{Mbata_2024}. The increasing availability of demographic data and the analysis of travel behavior patterns have transformed the way planners approach infrastructure development \cite{Aditjandra_2013,Roy_2021}. Transportation planners are currently trying to make informed decisions about infrastructure placement and capacity requirements using these demographic data and travel behavior patterns. This move has allowed fairly precise predictions of population movement and transportation needs \cite{Muller_2024}. 

Despite advances in transportation engineering \cite{kyem2024pavecap,kyem2025context} and planning \cite{HARIPAVAN2023128,lechtenberg2025guiding}, cities continue to face significant challenges in infrastructure development and management. A critical issue is the existing infrastructure wastage, where functional structures must be demolished to accommodate expanding transportation networks \cite{Xu_2019,Ling_2022}. This demolition and reconstruction represent substantial financial costs and create environmental impacts and community disruption. A historical example is the construction of Interstate 95 in Miami in the 1960s, which flattened large portions of Overtown, a predominantly Black neighborhood, forcing approximately 10,000 residents to leave their homes \cite{Dewey_2020}. Such displacement issues are common in rapidly growing urban areas, where the pace of expansion can outweigh even careful planning efforts based on demographic projections \cite{Angel_2023}. Traditional planning approaches, while incorporating available data, often struggle to accurately predict long-term urban growth patterns and transportation needs across multiple decades. This limitation frequently results in infrastructure placement decisions that, though meeting immediate needs, become obstacles to future development and require costly modifications or complete replacement \cite{Mierzejewski_1998}. The challenge is further complicated by changing travel behaviors and settlement patterns, which can rapidly transform transportation demands in ways that historical data failed to anticipate.

Machine learning has emerged as an innovative approach to address infrastructure wastage challenges in transportation planning \cite{Bhartiya_2024, danyo2024improved, aboah2023real,aboah2023deepsegmenter,duah2024divneds}. Incorporating machine learning has improved our ability to predict traffic patterns \cite{Rasulmukhamedov_2024}, population growth, and infrastructure requirements with greater precision than traditional forecasting methods \cite{Grossman_2022}. Although machine learning algorithms can process vast amounts of historical demographic data, and travel patterns to identify trends that human analysts might miss \cite{Son_2023,danyo2024improved,aboah2019investigation}, this data is low-dimensional and cannot capture the full complexity needed for effective infrastructure and transportation planning. Furthermore, current algorithms cannot transform this low-dimensional demographic and behavioral data into high-dimensional spatial representations that planners need. As a result, transportation planners cannot yet predict how these changes will manifest in the physical urban landscape. This limitation often leads to suboptimal decision making in infrastructure planning and design, as transportation planners struggle to fully understand how their decisions will impact the physical urban environment and existing communities over time. Predicting future urban settings would be valuable in preventing situations similar to the Overtown displacement, where the full spatial and social implications of infrastructure decisions were not adequately considered.

Our research addresses this gap by developing an integrated algorithm that not only predicts demographics and travel behavior but also predicts (generates) satellite imagery of future urban settings. Our novel approach combines demographic forecasting with Generative Adversarial Networks (GANs) to provide planners with visual representations of potential urban developments. Our method addresses the potential challenges that come with urban transportation planning, including the likelihood of premature demolition and future infrastructure wastage. This integration of predictive analytics with spatial visualization will transform infrastructure planning into a more proactive process, leading to more sustainable urban development and minimizing the need for costly infrastructure modifications in the future \cite{10807178}.

To address these limitations, this paper makes the following key contributions:

\begin{enumerate} 
    \item We propose a novel approach for injecting low-dimensional tabular data into a generative adversarial network (GAN) to produce high-quality satellite imagery, bridging the gap between demographic trends and spatial visualization. 
    \item We develop an integrated system that allows transportation planners to forecast urban transformations by leveraging demographic and travel behavior data, reducing the likelihood of infrastructure wastage and unplanned urban sprawl. 
    \item Our approach enables a proactive planning strategy by combining temporal forecasting models with generative image synthesis, offering a comprehensive tool for predicting the long-term spatial impact of transportation decisions. 
\end{enumerate}

The remainder of the paper is structured as follows. Section~\ref{sec:related_work} discusses related studies on transportation forecasting and urban planning using machine learning. Section~\ref{sec:data} talks about the data used and how it was obtained. Section~\ref{sec:methodology} presents our proposed framework, detailing the integration of demographic forecasting with satellite image generation. Section~\ref{sec:experiments} describes the experimental setup, evaluation metrics, and baseline comparisons. Section~\ref{sec:results} presents the results and analysis of our approach. Finally, Section~\ref{sec:conclusion} provides concluding remarks and directions for future research.

\section{Related Work}\label{sec:related_work}
\subsection{Historical Challenges in Transportation Infrastructure Planning and Their Socio-Spatial Consequences}
Planning for transport infrastructure has historically faced numerous challenges that have led to significant socio-spatial consequences in urban environments \cite{Serdar_2022}. Though some of these transportation infrastructures have aimed at improving mobility, these decisions have often led to unintended consequences such as community displacement, social segregation, and the creation of physical barriers within cities \cite{PAPADAKIS2024100125}. In the post-World War II era, many cities in Europe embraced auto-centric development patterns that resulted in extensive highway systems cutting through established neighborhoods. This transformation was obvious in cities like Rotterdam, which used post-war reconstruction as an opportunity to implement modernist planning principles favoring automobile circulation. This phenomenon resulted in a critical issue during this period, where infrastructure decisions were made based on contemporary needs without adequate consideration of future urban growth patterns and evolving transportation technologies \cite{Nuissl_2021}. This short-sighted approach has resulted in numerous cases of premature infrastructure obsolescence across cities worldwide. Boston's Central Artery, an elevated highway completed in 1959, became a classic example of infrastructure wastage due to urban expansion and the need for modernized transportation systems \cite{NAS_2003}. By the late 20th century, the highway was functionally obsolete, carrying over 200,000 vehicles daily—far exceeding its original capacity of 75,000—and causing severe congestion, high accident rates, and economic inefficiencies. It also physically divided neighborhoods such as the North End and the Waterfront from downtown Boston, limiting their economic potential \cite{NAE_2003}. The Boston Central Artery example and others have highlighted the limitations of historical planning approaches by recent advances in urban planning methodologies. The emergence of data-driven planning techniques, big data and sophisticated modeling tools \cite{10807178, danyo2024improved} has revealed infrastructure inefficiency and under-use patterns that could have been avoided with better predictive capabilities \cite{Perez-Martinez_2023,Vasilieva_2024,owor2023image2pci,owor2024pavesam}. Cities are now increasingly turning to predictive modeling and scenario planning to avoid the costly mistakes of the past. These modern approaches incorporate multiple data sources, including demographic trends, travel behavior patterns, and satellite imagery analysis, to better anticipate future urban development patterns and infrastructure needs \cite{Yu_2023, 10807178,aboah2023ai,behzadian20221st}.

\subsection{Challenges in Predicting High-Dimensional Spatial Outcomes from Low-Dimensional Data}
Urban planners often face limitations during urban transportation planning when projecting future urban infrastructural needs based on current socio-demographic data and travel patterns alone. Notable among these limitations in transportation planning is predicting high-dimensional spatial representations using low-dimensional input data. This limitation is compounded by the when sparse high-dimensional data reduces model accuracy \cite{Peng_Gui_Wu_2023, El-Sheikh_Ali_Abonazel_2024,aboah2020smartphone,wang2023gazesam}, and by redundancy in features that skew variance toward a few dimensions, masking critical urban complexity \cite{Peng_Gui_Wu_2023, Keningson_2024}. Traditional forecasting methods often struggle to capture spatial relationships and interdependencies that characterize urban settings. \cite{Manson_2007, Koldasbayeva_Tregubova_Gasanov_Zaytsev_Petrovskaia_Burnaev_2023}. Manson \cite{Manson_2007} highlights how methodological challenges in scaling and conflating patterns with processes create a disconnect between model outputs and real-world geographic dynamics. Urban systems also exhibit nonlinear interactions between demographics, transportation, and physical form, which low-dimensional data cannot adequately represent over time \cite{Auddy_Xia_Yuan_2024,Chevance_2024}. Lack of consistent, well-organized, and long-term data on how people move and travel over time in certain areas and contexts, further limits planners’ ability to predict evolving infrastructure needs \cite{Chevance_2024}. Emerging techniques like topological data analysis and big data methods offer promising solutions by modeling complex data structures with improved interpretability \cite{Auddy_Xia_Yuan_2024,Keningson_2024}, though challenges such as computational demands and data scalability must be addressed \cite{10807178, kyem2024advancingpavementdistressdetection}.

\section{Data}\label{sec:data}
Our study integrates demographic data and travel data from the U.S. Census Bureau, specifically from the American Community Survey (ACS), spanning 2012 to 2023 across approximately 58,000 California census tracts. This dataset includes 14 features covering demographics, modal split characteristics, and emerging travel trends. Additionally, high-resolution satellite imagery from the Mapbox API provides spatial context, enabling a comprehensive analysis of the interplay between socio-demographics, transportation infrastructure, and mobility patterns over time. A brief summary of our dataset is described in Table ~\ref{tab:table1}.
\begin{table}[t]
\caption{A summary of features in dataset}
\label{tab:table1}
\renewcommand{\arraystretch}{1.2}  
\begin{tabular}{p{0.35\columnwidth}p{0.55\columnwidth}}  
\hline
\textbf{Data Category} & \textbf{Feature} \\ \hline
\multirow{8}{*}{\parbox{0.35\columnwidth}{Demographic Data}} 
    & Total population \\
    & Percentage of people aged 25-34 \\
    & Percentage of people aged 35-50 \\
    & Percentage of people over 65 \\
    & Percentage of white population \\
    & Percentage of non-white population \\
    & Percentage of black population \\
    & Percentage with college degree \\
     & Average Income per capita \\
    \hline
\multirow{6}{*}{\parbox{0.35\columnwidth}{Travel Behavior Data}} 
    & Average travel time to work \\
    & Number of automobile users \\
    & Number of active transportation users (walking, cycling) \\
    & Number of public transit users \\
    & Number of other transportation mode users \\ \hline
\end{tabular}
\end{table}

\section{Methodology}\label{sec:methodology}
Our research is focused on developing a model that uses low-dimensional demographic and travel behavior data to generate satellite imagery of future urban settings. To achieve this, we propose a two-stage framework which uses a temporal fusion transformer to forecast travel behavior based on demographic data, capturing the underlying patterns and dynamics over several years. The outputs from this stage is then used as input for a modified StyleGAN 2 model, which then generates synthetic satellite images to predict how the area might evolve. This innovative approach combines state-of-the-art time-series analysis with advanced image generation, offering a comprehensive perspective on urban development and spatial change.

\subsection{Data Preprocessing}
\noindent
The dataset was first examined to remove data entries with zeros, outliers were then identified using the interquartile range (IQR), calculated as
\begin{equation}
   \mathrm{IQR} = Q_{3}(X) - Q_{1}(X), 
\end{equation}

where \(Q_{1}(X)\) and \(Q_{3}(X)\) denote the first and third quartiles of the feature set \(X\), respectively. Any observation \(x\) lying outside 
\(\bigl[\,Q_{1}(X) - 1.5 \cdot \mathrm{IQR},\,Q_{3}(X) + 1.5 \cdot \mathrm{IQR}\bigr]\)
was discarded if deemed erroneous. Finally, numeric variables were standardized via z-score scaling,

\begin{equation}
    Z(x) = \frac{x - \mu}{\sigma},
\end{equation}

where \(\mu\) is the mean value of the feature \(x\), and \(\sigma\) is its standard deviation, to ensure uniform feature ranges for subsequent modeling.

\noindent Following preprocessing, the dataset was partitioned chronologically into training and testing sets. The training set comprised approximately 67\% of observations (approximately 39,000 entries) from the period 2012-2017, while the testing set contained the remaining 33\% ( which is about 19,000 entries) from 2018-2023.

\begin{align}
\mathcal{D}_{\text{train}} &= \{(x_i, y_i) \in \mathcal{D} \mid 2012 \leq t_i \leq 2017\} \\
\mathcal{D}_{\text{test}} &= \{(x_i, y_i) \in \mathcal{D} \mid 2018 \leq t_i \leq 2023\}
\end{align}

\noindent This temporal division ensures the model trains on historical data and evaluates on future observations, simulating real-world forecasting conditions.

\subsection{Temporal Fusion Transformer for Travel Behavior Prediction}

The Temporal Fusion Transformer (TFT) serves as the first stage of our predictive framework, where it learns historical relationships between demographic features and travel behavior patterns. The goal of this stage is to forecast transportation mode distributions based on past and present population dynamics, socioeconomic attributes, and racial composition. The output of the TFT is then used as a conditioning input for the Generative Adversarial Network (GAN) to generate high-resolution satellite imagery that visualizes predicted spatial transformations.

\subsubsection{Problem Formulation}

Let $\mathbf{X} \in \mathbb{R}^{T \times d}$ denote the time series of demographic features, where $T$ represents the number of past time steps, and $d$ is the number of demographic input features. Our objective is to predict future travel behavior trends $\mathbf{Y} \in \mathbb{R}^{T' \times k}$, where $T'$ is the forecasting horizon and $k$ represents the number of target transportation variables.

The input features (demographic data) are defined as:
\begin{equation}
    \mathbf{X} = \{ x_t \}_{t=1}^{T}, \quad x_t \in \mathbb{R}^{d}
\end{equation}

The output features (travel behavior data) are defined as:
\begin{equation}
    \mathbf{Y} = \{ y_t \}_{t=T+1}^{T+T'}, \quad y_t \in \mathbb{R}^{k}
\end{equation}

\subsubsection{Encoder-Decoder Architecture}

The TFT employs an encoder-decoder framework to model sequential dependencies between demographic features and future travel behavior. The encoder processes historical demographic data into a latent representation, which the decoder then uses to make multi-step forecasts.

The encoder uses a Long Short-Term Memory (LSTM) network to capture long-term dependencies:
\begin{equation}
    \mathbf{h}_t = \text{LSTM}(\mathbf{x}_t, \mathbf{h}_{t-1}, \mathbf{c}_{t-1})
\end{equation}

where, $\mathbf{h}_t$ is the hidden state at time step $t$, $\mathbf{c}_t$ is the cell state of the LSTM and $\mathbf{x}_t$ represents the demographic input at time $t$.

The decoder then takes the encoded representation and generates travel behavior predictions for the future time steps:

\begin{equation}
    \hat{\mathbf{y}}_{t+1} = \text{Decoder}(\mathbf{h}_T)
\end{equation}

where, $\hat{\mathbf{y}}_{t+1} \in \mathbb{R}^{k}$ represents the predicted travel behavior variables at future time step $t+1$, $\mathbf{h}_T \in \mathbb{R}^{m}$ is the final hidden state output from the LSTM encoder, summarizing all past demographic data up to time step $T$, $\text{Decoder}(\cdot)$ is a learned function that maps the encoded demographic representation $\mathbf{h}_T$ to travel behavior predictions, using attention mechanisms and fully connected layers, $k$ is the number of travel behavior features (e.g., number of automobile users, transit users, etc.), $m$ is the dimension of the encoded hidden state, which captures temporal dependencies in the demographic input.

\subsubsection{Multi-Head Attention Mechanism}
To model the temporal dependencies between demographic data and future travel behavior, we employ a multi-head attention mechanism that selectively focuses on the most relevant historical patterns. This allows the model to assign varying importance to past demographic attributes when forecasting transportation trends.

At each time step, the query, key, and value matrices are defined as:
\begin{equation}
    \mathbf{Q}, \mathbf{K}, \mathbf{V} \in \mathbb{R}^{T \times d}
\end{equation}
where, $\mathbf{Q}$ represents the current demographic state influencing travel behavior, $\mathbf{K}$ encodes past demographic trends and $\mathbf{V}$ contains the corresponding travel behavior information.

The scaled dot-product attention is computed as:
\begin{equation}
    \text{Attention}(\mathbf{Q}, \mathbf{K}, \mathbf{V}) = \text{softmax} \left( \frac{\mathbf{Q} \mathbf{K}^T}{\sqrt{d}} \right) \mathbf{V}
\end{equation}
where the softmax function ensures that weights are normalized, enhancing the most relevant historical influences.

To capture multiple perspectives in the data, we apply multi-head attention, where multiple attention mechanisms operate in parallel:

\begin{equation}
    \text{MultiHead}(\mathbf{Q}, \mathbf{K}, \mathbf{V}) = \text{Concat}(\text{head}_1, \text{head}_2, \dots, \text{head}_h) \mathbf{W}^O
\end{equation}
where each attention head independently learns different temporal relationships between demographics and travel behavior and $\mathbf{W}^O$ is a trainable projection matrix ensuring dimensional consistency.

Through this mechanism, our model is able to dynamically attend to the most influential time steps, enabling the TFT to accurately model how past demographic changes impact future travel behavior.

\subsubsection{Feed-Forward Network}

After processing historical demographic trends through attention and sequential modeling, the feed-forward network (FFN) generates the final travel behavior predictions by refining the extracted temporal representations.

At each time step, the hidden state $\mathbf{h}_t$ undergoes a nonlinear transformation:

\begin{equation}
    \mathbf{z}_t = \sigma(\mathbf{W}_1 \mathbf{h}_t + \mathbf{b}_1)
\end{equation}
\begin{equation}
    \mathbf{h}_t^{'} = \mathbf{W}_2 \mathbf{z}_t + \mathbf{b}_2
\end{equation}

where $\sigma(\cdot)$ represents the ReLU activation function, introducing non-linearity to capture complex relationships between demographics and mobility patterns and $\mathbf{W}_1, \mathbf{W}_2, \mathbf{b}_1, \mathbf{b}_2$ are learnable parameters that optimize feature extraction.

The final predictions, representing future travel behavior metrics, are computed as:

\begin{equation}
    \hat{y}_t = \mathbf{W}_{out} \mathbf{h}_t^{'} + \mathbf{b}_{out}
\end{equation}

where $\mathbf{W}_{out}$ and $\mathbf{b}_{out}$ project the refined hidden states ($\mathbf{h}_t^{'}$) into the target travel behavior space.

\subsubsection{Loss Function and Optimization}

To ensure robust and stable training, we employ the smooth L1 loss, which balances sensitivity to small deviations while controlling the influence of outliers in travel behavior predictions:

\begin{equation}
    L(\hat{y}, y) = \sum_t 
    \begin{cases} 
      0.5 (\hat{y}_t - y_t)^2, & \text{if } |\hat{y}_t - y_t| < 1 \\
      |\hat{y}_t - y_t| - 0.5, & \text{otherwise}
    \end{cases}
\end{equation}

where $\hat{y}_t$ and $y_t$ represent the predicted and actual travel behavior metrics, respectively. The quadratic term penalizes small errors, ensuring precise predictions for stable trends and the linear term prevents instability from large deviations, making the model robust to outliers.

To optimize model performance, we utilize the Adam optimizer, which dynamically adjusts learning rates for each parameter, improving convergence speed and stability:
\begin{equation}
    \theta_{t+1} = \theta_t - \eta \cdot \frac{m_t}{\sqrt{v_t} + \epsilon}
\end{equation}
where $\eta$ is the learning rate, adapting to gradient variations over time. $m_t$ and $v_t$ also represent the first and second moment estimates, stabilizing updates and preventing vanishing or exploding gradients.

Combining temporal modeling, attention mechanisms, and robust optimization, the TFT effectively forecasts future travel behavior trends based on evolving demographic characteristics, ensuring reliable inputs for the subsequent GAN-based spatial prediction.

\subsection{Generative Adversarial Network for Spatial Prediction}

The second stage of our model employs a Generative Adversarial Network (GAN) to synthesize realistic satellite images that reflect predicted spatial changes based on travel behavior. Unlike conventional GANs that rely on latent noise vectors or text prompts, our approach integrates structured tabular data representing mobility trends. This enables the model to generate data-driven urban landscapes, ensuring interpretability grounded in real-world transportation patterns.

\subsubsection{Tabular Data Injection}
The GAN receives structured tabular data as input, encoding travel behavior patterns across different regions and time periods. Given an input dataset $\mathbf{X} \in \mathbb{R}^{N \times d}$, where $N$ is the number of regions and $d$ represents features such as average travel time to work, population of automobile users, active transport users (walking and cycling), public transit users and other transport mode users, we transform these into a latent representation while preserving spatial correlations. To achieve this, the travel behavior data is concatenated with a randomly sampled latent vector $\mathbf{z} \sim \mathcal{N}(0, I)$:
\begin{equation}
    \mathbf{w} = f_{\text{enc}}([\mathbf{X}, \mathbf{z}])
\end{equation}
where $f_{\text{enc}}(\cdot)$ is a trainable function mapping the concatenated input to a high-dimensional space, allowing travel behavioral patterns to interact with the generative process.

\begin{figure}[h]
\centering
\includegraphics[scale=0.45]{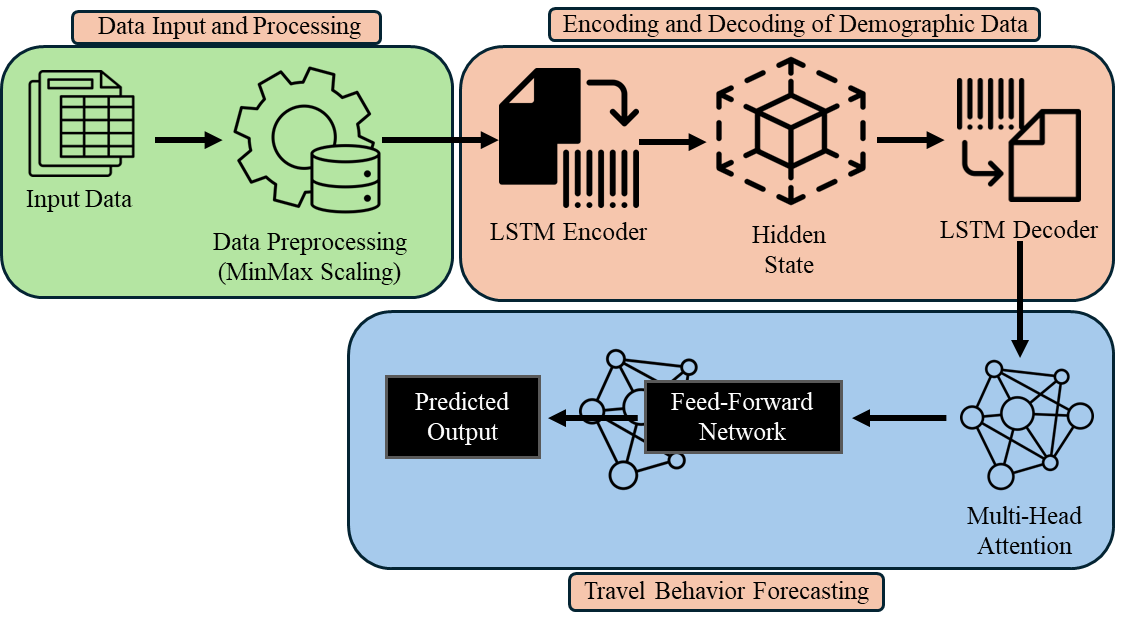}
\caption{ Overview of the proposed framework for the Temporal Fusion Transformer}
\end{figure}

\subsubsection{Generator Architecture}

The generator $G$ learns to produce satellite images conditioned on travel behavior data:

\begin{equation}
    \mathbf{I}_{\text{fake}} = G(\mathbf{w})
\end{equation}

where $\mathbf{I}_{\text{fake}}$ is the generated satellite image. To prevent loss of travel behavior information, we incorporate Adaptive Instance Normalization (AdaIN):

\begin{equation}
    \text{AdaIN}(\mathbf{h}, \mathbf{w}) = \gamma(\mathbf{w}) \cdot \frac{\mathbf{h} - \mu(\mathbf{h})}{\sigma(\mathbf{h})} + \beta(\mathbf{w})
\end{equation}

where $\mathbf{h}$ represents intermediate feature maps, while $\gamma(\mathbf{w})$ and $\beta(\mathbf{w})$ adaptively scale and shift them based on the tabular input.

Additionally, Noise Injection is applied at each resolution to introduce controlled stochasticity, ensuring diversity in generated satellite images:

\begin{equation}
    \mathbf{h}' = \mathbf{h} + \alpha \cdot \mathbf{N}, \quad \mathbf{N} \sim \mathcal{N}(0, I)
\end{equation}

where $\alpha$ is a learnable weight controlling noise intensity.

\subsubsection{Discriminator Architecture}

The discriminator, denoted as $D$, is a crucial component of our GAN framework, responsible for evaluating the realism of generated satellite images while ensuring their consistency with the expected travel behavior trends. It serves as a binary classifier, distinguishing between real satellite images $\mathbf{I}_{\text{real}}$ and synthetic images $\mathbf{I}_{\text{fake}}$ produced by the generator, conditioned on tabular travel behavior data $\mathbf{X}$. This process can be mathematically expressed as:

\begin{equation}
    D(\mathbf{I}, \mathbf{X}) \to y
\end{equation}

where, $\mathbf{I} \in \mathbb{R}^{H \times W \times C}$ represents the satellite image (real or generated), where $H$ and $W$ are spatial dimensions and $C$ is the number of channels, $\mathbf{X} \in \mathbb{R}^{d}$ is the corresponding travel behavior data and $y \in [0,1]$ is a scalar output, representing the probability that $\mathbf{I}$ is a real image.
\begin{figure}[ht]
\centering
\includegraphics[scale=0.43]{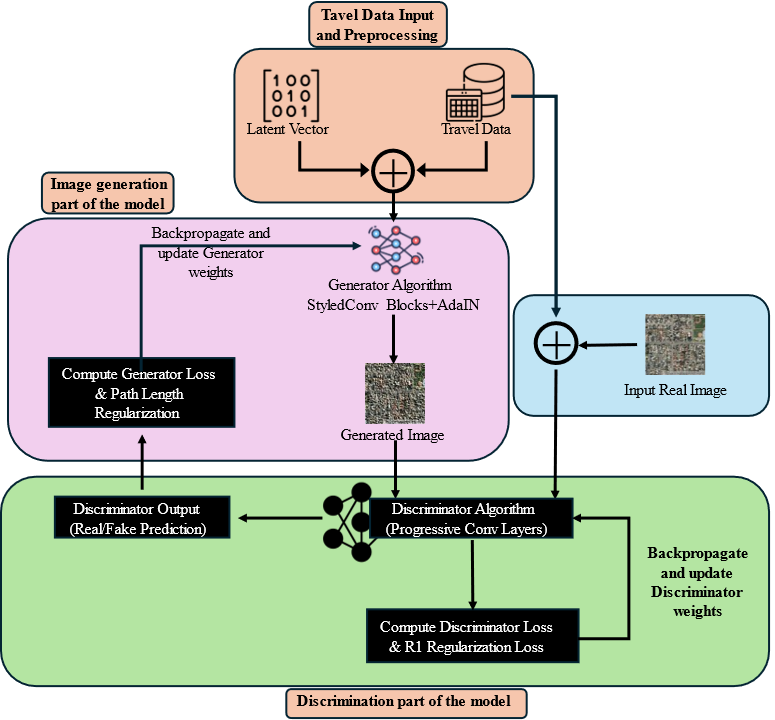}
\caption{ Overview of the proposed framework for the Generative Adversarial Network}
\end{figure}
Unlike traditional discriminators that solely process image inputs, our discriminator jointly evaluates satellite imagery and travel behavior data. This ensures that generated images do not just appear realistic but also adhere to real-world transportation trends. The conditioning process is implemented via feature fusion:

\begin{equation}
    \mathbf{h}_D = f_{\text{img}}(\mathbf{I}) + f_{\text{tab}}(\mathbf{X})
\end{equation}

where, $f_{\text{img}}(\cdot)$ extracts spatial features from the image using convolutional layers, $f_{\text{tab}}(\cdot)$ processes tabular features through a multi-layer perceptron (MLP) and the feature representations are fused through element-wise addition before classification.

A common issue in GAN training is mode collapse, where the generator produces limited variations of images, reducing diversity. To address this, we integrate a Minibatch Standard Deviation (Minibatch-STD) layer, which enhances generalization by encouraging the generator to produce diverse and realistic outputs.

This is achieved by computing the statistical variation within each batch of generated images:

\begin{equation}
    \mathbf{h}' = \left[ \mathbf{h}, \text{std}(\mathbf{h}) \right]
\end{equation}

where, $\mathbf{h}$ represents the intermediate feature map of the discriminator, $\text{std}(\mathbf{h})$ computes the per-channel standard deviation across the minibatch, the computed standard deviation is concatenated with $\mathbf{h}$, allowing the discriminator to detect whether an image lacks diversity. The final output is produced by passing the combined features through fully connected layers, leading to a single probability score:

\begin{equation}
    y = \sigma(\mathbf{W}_D \mathbf{h}' + \mathbf{b}_D)
\end{equation}

where, $\mathbf{W}_D$ and $\mathbf{b}_D$ are trainable weights and biases of the final classification layer and $\sigma(\cdot)$ is the sigmoid activation function, ensuring the output probability is within $[0,1]$.

This structured approach allows the discriminator to enforce both visual realism and transportation consistency, ensuring that generated satellite images align with predicted travel behavior trends.

\subsubsection{Loss Functions}

The training of our GAN follows a non-saturating adversarial loss formulation, ensuring effective gradient flow during optimization. Additionally, a gradient penalty is introduced to stabilize training and enforce Lipschitz continuity, reducing mode collapse and improving overall sample quality.

\paragraph{Discriminator Loss}
The discriminator is trained to distinguish between real satellite images $\mathbf{I}_{\text{real}}$ and generated images $\mathbf{I}_{\text{fake}}$ while ensuring that the generated outputs align with real-world travel behavior trends. The loss function for the discriminator, $L_D$, consists of two terms:

\begin{equation}
    L_D = \mathbb{E} \left[ \log D(\mathbf{I}_{\text{real}}, \mathbf{X}) \right] + \mathbb{E} \left[ \log (1 - D(G(\mathbf{w}), \mathbf{X})) \right]
\end{equation}

where $D(\mathbf{I}_{\text{real}}, \mathbf{X})$ represents the probability that a real satellite image $\mathbf{I}_{\text{real}}$ aligns with the corresponding travel behavior data $\mathbf{X}$, $D(G(\mathbf{w}), \mathbf{X})$ represents the probability that the generated image $\mathbf{I}_{\text{fake}} = G(\mathbf{w})$ is real. The discriminator maximizes this loss by increasing its confidence in classifying real images as real and fake images as fake.

\paragraph{Generator Loss}
The generator is trained to fool the discriminator into classifying its outputs as real, meaning it aims to maximize the discriminator’s probability of labeling generated images as real. This is achieved by minimizing the following objective:

\begin{equation}
    L_G = - \mathbb{E} \left[ \log D(G(\mathbf{w}), \mathbf{X}) \right]
\end{equation}

The generator seeks to maximize $D(G(\mathbf{w}), \mathbf{X})$. This formulation encourages the generator to produce satellite images that not only appear realistic but also correspond accurately to mobility patterns.

\paragraph{Gradient Penalty for Stability}
To improve training stability and enforce Lipschitz continuity, we incorporate a gradient penalty inspired by Wasserstein GAN with Gradient Penalty (WGAN-GP). This regularization term ensures that the discriminator maintains smooth gradients, reducing the risk of vanishing or exploding gradients. The gradient penalty is defined as:

\begin{equation}
    L_{\text{GP}} = \lambda \mathbb{E} \left[ \left( \|\nabla_{\mathbf{\hat{I}}} D(\mathbf{\hat{I}}, \mathbf{X}) \|_2 - 1 \right)^2 \right]
\end{equation}

where $\mathbf{\hat{I}}$ is an interpolated sample between a real and a generated image, ensuring smooth transitions in feature space, $\lambda$ is a hyperparameter controlling the strength of the gradient penalty. The penalty encourages gradients to have a unit norm, preventing sharp discriminator updates that could destabilize training.

The final optimization problem for the discriminator and generator is formulated as:

\begin{equation}
    \begin{split}
        \min_G \max_D \quad \mathbb{E} \left[ \log D(\mathbf{I}_{\text{real}}, \mathbf{X}) \right] + \\ \mathbb{E}
        \left[ \log (1 - D(G(\mathbf{w}), \mathbf{X})) \right] + \lambda L_{\text{GP}}
    \end{split}
\end{equation}

This adversarial training process ensures that:
\begin{itemize}
    \item The discriminator becomes better at distinguishing real satellite images from generated ones.
    \item The generator continuously improves, producing high-fidelity spatial predictions that align with forecasted travel behavior.
    \item The gradient penalty stabilizes training, preventing sudden divergence or mode collapse.
\end{itemize}

Through the combination of adversarial learning with gradient regularization, our model ensures that generated satellite images accurately reflect predicted transportation trends, making the outputs both visually realistic and empirically interpretable.

\section{Evaluation Metrics}
To assess the performance of our proposed model, we evaluate both the Temporal Fusion Transformer (TFT) for travel behavior forecasting and the Generative Adversarial Network (GAN) for urban setting prediction. The evaluation framework consists of statistical accuracy metrics for the TFT and perceptual similarity metrics for the GAN.

\subsection{Temporal Forecasting Evaluation}

The TFT is evaluated using three key time-series regression metrics: Root Mean Square Error (RMSE), R-Squared ($R^2$), and Dynamic Time Warping (DTW).

\subsubsection{Root Mean Square Error (RMSE)}
RMSE measures the standard deviation of prediction errors, penalizing large deviations more heavily than other error metrics:

\begin{equation}
    \text{RMSE} = \sqrt{\frac{1}{T} \sum_{t=1}^{T} (\hat{y}_t - y_t)^2}
\end{equation}

where $\hat{y}_t$ and $y_t$ denote the predicted and actual travel behavior metrics at time $t$, respectively.

\subsubsection{R-Squared ($R^2$ Score)}

The $R^2$ score evaluates how well the model explains the variance in the actual data:

\begin{equation}
    R^2 = 1 - \frac{\sum_{t=1}^{T} (\hat{y}_t - y_t)^2}{\sum_{t=1}^{T} (y_t - \bar{y})^2}
\end{equation}

where $\bar{y}$ represents the mean of the actual values. A score close to $1$ indicates that the model effectively captures underlying trends in travel behavior.

\subsubsection{Dynamic Time Warping (DTW)}

Since travel behavior data often exhibits temporal shifts, DTW is used to measure similarity between the predicted and actual sequences:

\begin{equation}
    \text{DTW}(\hat{\mathbf{y}}, \mathbf{y}) = \min_{\pi} \sum_{(i,j) \in \pi} D_{i,j}
\end{equation}

where $\pi$ represents an optimal alignment path, and $D_{i,j}$ is the cost of aligning $\hat{y}_i$ with $y_j$. Lower DTW scores indicate better alignment between the predicted and actual time series.

\subsection{Spatial Prediction Evaluation}

The GAN-generated satellite images are evaluated using Fréchet Inception Distance (FID) and Structural Similarity Index Measure (SSIM) to assess realism and consistency with real-world satellite imagery.

\subsubsection{Fréchet Inception Distance (FID)}

FID quantifies the perceptual similarity between real and generated satellite images by comparing their feature distributions:

\begin{equation}
    \text{FID} = \|\mu_r - \mu_g\|^2 + \text{Tr}(\Sigma_r + \Sigma_g - 2 (\Sigma_r \Sigma_g)^{1/2})
\end{equation}

where $(\mu_r, \Sigma_r)$ and $(\mu_g, \Sigma_g)$ are the mean and covariance matrices of real and generated images in a deep feature space. Lower FID scores indicate more realistic image generation.

\subsubsection{Structural Similarity Index Measure (SSIM)}

SSIM evaluates the perceptual quality of generated images by measuring luminance, contrast, and structural similarities:

\begin{equation}
    \text{SSIM}(\mathbf{I}_{\text{real}}, \mathbf{I}_{\text{fake}}) = \frac{(2\mu_r \mu_g + C_1)(2\sigma_{rg} + C_2)}{(\mu_r^2 + \mu_g^2 + C_1)(\sigma_r^2 + \sigma_g^2 + C_2)}
\end{equation}

where $\mu_r$, $\mu_g$ are the mean pixel values, $\sigma_r^2$, $\sigma_g^2$ are variances, and $\sigma_{rg}$ is the covariance. SSIM values range from $0$ to $1$, with higher values indicating stronger perceptual similarity.

\section{Experiments}\label{sec:experiments}

To evaluate the effectiveness of our proposed framework, we designed two sets of experiments: (1) Travel Behavior Prediction, where we compare the Temporal Fusion Transformer (TFT) against baseline time-series models, and (2) Spatial Prediction, where we assess the impact of varying the latent space in a Generative Adversarial Network (GAN).  
All experiments were conducted on an NVIDIA A40 GPU with 32 GB of memory. 

\subsection{Travel Behavior Prediction}

For forecasting future travel behavior, we trained four models: RNN, LSTM, LSTM with Attention, and TFT.  

Each model was trained to predict the travel behavior based on historical demographic trends. The input sequence length was set to 3 years, forecasting the next 3 years.

\subsubsection{Model Configurations and Training}
The hyperparameters used for all models are summarized in Table~\ref{tab:time_series_models}.  
\begin{table}[ht]
    \centering
    \caption{Hyperparameters for Travel Behavior Forecasting Models}
    \label{tab:time_series_models}
    \renewcommand{\arraystretch}{1.2}
    \scalebox{0.7}{
        \begin{tabular}{lccccccc}
            \hline
            \textbf{Model} & \textbf{Hidden Size} & \textbf{Layers} & \textbf{Dropout} & \textbf{Attention} & \textbf{Epochs} & \textbf{Batch} \\
            \hline
            RNN & 128 & 2 & 0.1 & None & 200 & 16 \\
            LSTM & 256 & 2 & 0.1 & None & 200 & 16 \\
            LSTM + Attention & 256 & 2 & 0.1 & Bahdanau & 200 & 16 \\
            TFT & 128 & 4 & 0.1 & Multi-Head & 200 & 16 \\
            \hline
        \end{tabular}}
\end{table}

All models were trained using the Adam optimizer with a learning rate of $5 \times 10^{-4}$ and a Smooth L1 Loss function.

The TFT was additionally optimized with multi-head self-attention, which assigns weights to important temporal patterns

\subsection{Spatial Prediction with GAN}
The second stage of our experiment evaluates GAN-based spatial prediction, where we generate synthetic satellite images of urban evolution based on predicted travel behavior. To achieve this we investigated how different latent space sizes affect image generation. The GAN generator was trained using three configurations:

\begin{table}[H]
    \centering
    \caption{Hyperparameters for GAN-based Spatial Prediction}
    \label{tab:gan_models}
    \renewcommand{\arraystretch}{1.2}
    \begin{tabular}{lccc}
        \hline
        \textbf{Model} & \textbf{Latent Dim} & \textbf{Learning Rate} & \textbf{Batch Size} \\
        \hline
        GAN-128 & 128 & $8 \times 10^{-5}$ & 16 \\
        GAN-256 & 256 & $8 \times 10^{-5}$ & 16 \\
        GAN-512 & 512 & $8 \times 10^{-5}$ & 16 \\
        \hline
    \end{tabular}
\end{table}

The GAN models were trained for 50,000 iterations using the Adam optimizer with separate learning rates:

\begin{itemize}
    \item Generator Learning Rate: $8 \times 10^{-5}$
    \item Discriminator Learning Rate: $3 \times 10^{-5}$
\end{itemize}

Regularization techniques included:
\begin{itemize}
    \item R1 Gradient Penalty: $\gamma = 10$
    \item Path Length Regularization: Weight = 2
\end{itemize}

\section{Results and Discussion}\label{sec:results}
In evaluating the effectiveness of the Temporal Fusion Transformer (TFT) against traditional time-series models, including LSTM, LSTM with Attention, and RNN, we analyzed forecasting accuracy using Root Mean Square Error (RMSE), R-Squared ($R^2$), and Dynamic Time Warping (DTW).

Table~\ref{tab:model_comparison} summarizes the performance metrics, demonstrating that TFT outperformed all baseline models by achieving the lowest RMSE and highest $R^2$ - indicating an improved predictive accuracy. The lower DTW distance further suggests that TFT captured temporal dependencies more effectively, aligning closely with actual travel behavior trends.

\begin{table}[ht]
    \centering
    \caption{Performance Comparison of Time-Series Models}
    \label{tab:model_comparison}
    \renewcommand{\arraystretch}{1.2}
    \begin{tabular}{lccc}
        \hline
        \textbf{Model} & \textbf{RMSE} $\downarrow$ & \textbf{$R^2$ Score} $\uparrow$ & \textbf{DTW Distance} $\downarrow$ \\
        \hline
        RNN & 8.72 & 0.61 & 15.3 \\
        LSTM & 7.94 & 0.68 & 12.7 \\
        LSTM + Attention & 7.45 & 0.72 & 10.9 \\
        TFT (Ours) & \textbf{7.28} & \textbf{0.76} & \textbf{10.2} \\
        \hline
    \end{tabular}
\end{table}

TFT’s better performance can be attributed to its multi-head attention mechanism, which effectively captures long-term dependencies and assigns dynamic feature importance. Unlike LSTM-based models, which rely solely on recurrent structures, TFT uses attention to weigh critical temporal relationships, leading to more stable and accurate forecasts.

Furthermore, while LSTM + Attention improves over standard LSTM by selectively focusing on important past data points, it lacks the fully integrated gating and feature selection mechanisms present in TFT, making it less robust for long-horizon forecasting.

\begin{figure}[ht]
    \centering
    \includegraphics[scale=0.32]{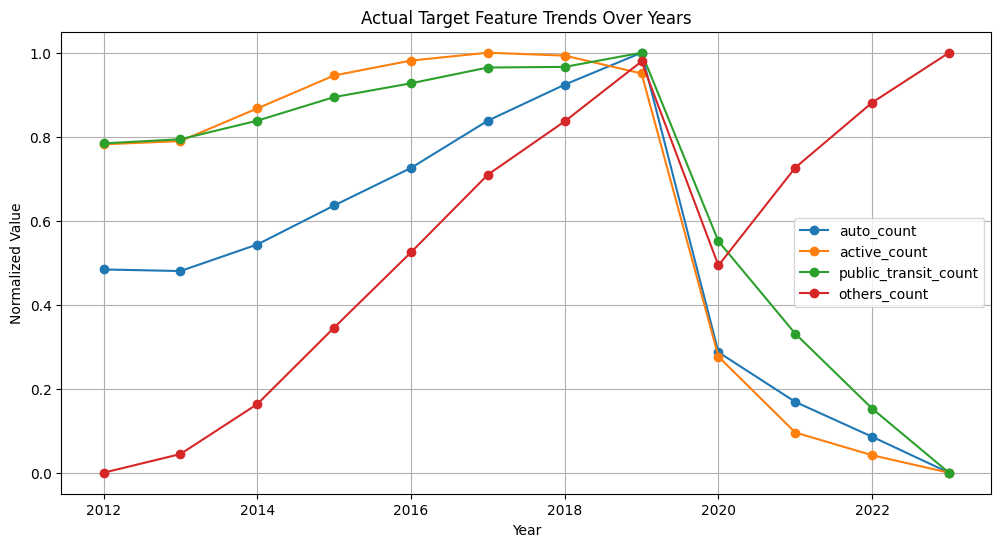}
    \caption{Actual Target (Travel Behavior) Feature Trends Over Years}
    \label{fig:actual_trends}
\end{figure}

\subsection{Discussion on Forecasting Performance}
\begin{figure}[ht]
    \centering
    \includegraphics[scale=0.5]{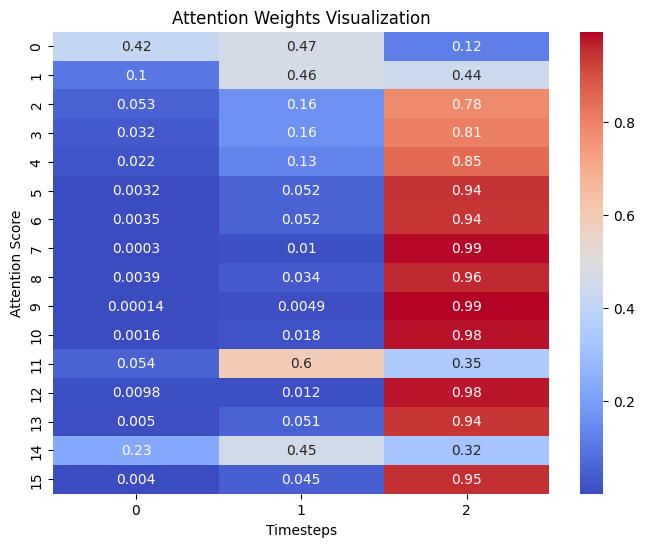}
    \caption{Attention Weights Visualization. The attention mechanism assigns varying importance to different time steps, but fails to emphasize the critical period leading up to 2020.}
    \label{fig:attention_weights}
\end{figure}

The LSTM model, designed to capture sequential dependencies, struggled to account for abrupt changes in travel behavior, particularly the sharp decline observed during the COVID-19 pandemic. The LSTM+Attention model aimed to improve upon this by dynamically weighting historical time steps. While the attention map (illustrated in Fig~\ref{fig:attention_weights}) indicates that the model assigned significant weight to the third timestep, which suggests an effort to capture critical historical patterns, it still failed to anticipate the drastic shift in mobility trends. Similarly, the RNN did not perform well, as it lacked the capability to effectively adapt to sudden disruptions in the data. In contrast, the TFT model demonstrated a better ability to follow overall trends, making it a more adaptable alternative. However, all models faced challenges in fully capturing sudden disruptions, highlighting the need for more robust architectures to handle extreme variations in temporal data.

One possible reason for this failure is that, although the model emphasized certain past trends, it still relied on patterns that followed normal fluctuations and did not adjust adequately to unforeseen external disruptions. This suggests that the attention mechanism was not sufficiently adaptive to shifts outside of its learned distributions. Future improvements could involve incorporating external variables, such as mobility restrictions or pandemic-related indicators, to allow the model to adjust more effectively to unexpected events.

\begin{figure}[ht]
    \centering
    \includegraphics[scale=0.32]{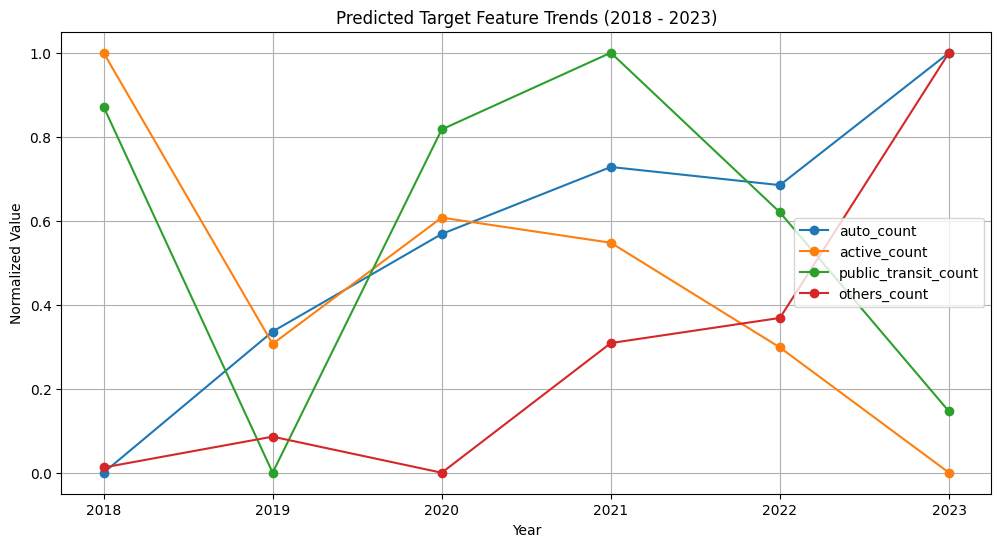}
    \caption{Predicted Target Feature Trends (2018 - 2023) using the TFT Model. The model successfully captures the overall trend in travel behavior, demonstrating its effectiveness.}
    \label{fig:tft_forecast}
\end{figure}

The Temporal Fusion Transformer demonstrated strong predictive capability, effectively capturing the overall trends in travel behavior, as seen in Figure~\ref{fig:tft_forecast}. Unlike the LSTM-based models, which struggled to adjust to abrupt changes, the RNN showed greater flexibility in modeling fluctuations in transportation trends. While some deviations from the actual values exist, the model was able to approximate key trend reversals, indicating its effectiveness in forecasting dynamic mobility patterns. The results suggest that despite its simplicity, the RNN remains a competitive option for travel behavior prediction, especially when trained on sufficient historical data.

These results validate the effectiveness of TFT in forecasting travel behavior, making it a strong candidate for mobility trend prediction. Next, we analyze feature importance and model interpretability.

To evaluate the impact of latent dimension size on image generation quality, we trained three variants of the GAN generator with latent dimensions of 128, 256, and 512. The Fréchet Inception Distance (FID) and Structural Similarity Index (SSIM) were used to measure image realism and alignment with real satellite images.

Table~\ref{tab:generator_comparison} summarizes the results, showing that the generator with a latent dimension of 512 performed best, achieving the lowest FID (indicating higher similarity to real images) and the highest SSIM (demonstrating better structural consistency).

\begin{table}[ht]
    \centering
    \caption{Generator Performance Comparison with Different Latent Dimensions}
    \label{tab:generator_comparison}
    \renewcommand{\arraystretch}{1.2}
    \begin{tabular}{lcc}
        \hline
        \textbf{Latent Dimension} & \textbf{FID Score} ↓ & \textbf{SSIM Score} ↑ \\
        \hline
        128  & 24.3 & 0.67 \\
        256  & 19.8 & 0.73 \\
        512  & \textbf{15.2} & \textbf{0.81} \\
        \hline
    \end{tabular}
\end{table}

The improved performance of larger latent dimensions suggests that a richer feature space enhances image quality, allowing the generator to capture finer spatial details and produce more realistic satellite images. While the 128-dimension generator struggled with image sharpness and structural accuracy, the 512-dimension generator consistently produced clearer, more accurate spatial representations.

\begin{figure}[ht]
    \centering
    \begin{minipage}{0.23\textwidth}
        \centering
        \includegraphics[width=\linewidth]{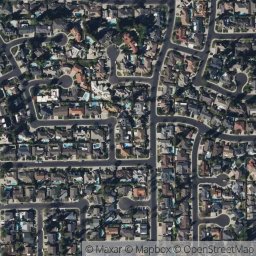}
    
        \label{fig:actual_satellite}
    \end{minipage}
    \hfill
    \begin{minipage}{0.23\textwidth}
        \centering
        \includegraphics[width=\linewidth]{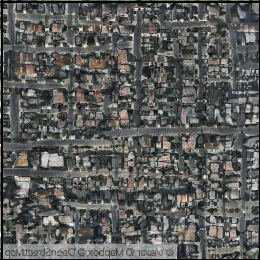}
        
        \label{fig:generated_satellite}
    \end{minipage}
    \caption{Comparison of Actual and Generated Satellite Images. (On the left is the actual image and on the right is the generated image)}
    \label{fig:gan_results}
\end{figure}

The proposed GAN effectively generates satellite images that exhibit structural similarity to real-world spatial layouts. As shown in Figure~\ref{fig:gan_results}, the generated image preserves key urban features, such as road networks and building distributions, making it a valuable tool for predicting urban evolution. However, slight distortions and artifacts suggest that further refinement is necessary for improved realism.

Our model demonstrated strong predictive capabilities. Trained on travel data from 2012 to 2017, it successfully forecasted 2021 travel behavior trends and generated high-quality satellite images, making it a valuable tool for transportation policy planning. However, some predictions deviated from actual travel patterns, highlighting potential areas for further refinement in long-term forecasting accuracy.

\subsection{Data Privacy}
\noindent
All demographic and travel data were used at the census-tract level, ensuring no personally identifiable information was involved. This aggregated approach preserves individual privacy while still capturing broad trends. Similarly, the satellite imagery, sourced from publicly available platforms, contains no personal identifiers. All datasets were stored on secure servers with restricted access, adhering to institutional privacy protocols and data usage agreements, thereby minimizing the risk of unauthorized disclosure.

\subsection{Practical Applicability of Study}
\subsubsection{Integration with Government Systems}
Since our framework utilizes aggregated publicly available demographic and travel behavior data along with high resolution satellite imagery, the study can be seamlessly integrated with existing government systems. These data streams, which are routinely collected and maintained by governmental agencies, can be directly incorporated into current Geographic Information Systems (GIS) and transportation management platforms. Such integration enables real-time scenario analyses and predictive forecasting of infrastructure needs, thereby facilitating proactive policy planning and urban development. The system’s adherence to stringent data privacy and security standards further supports its incorporation into established governmental workflows.

\subsubsection{Potential Applications}
Beyond governmental integration, this study offers a wide range of potential applications in urban planning and transportation management. The combined predictive analytics and spatial visualization capabilities not only enhance long-term infrastructure planning but also provide critical insights for immediate decision-making. 
Adaptable to real-time operations, the framework can support dynamic urban management by continuously updating predictions as new data become available. This real-time adaptability is particularly beneficial for responding to abrupt changes in travel behavior or unexpected urban events. Additionally, the approach can be extended to support scenario planning and predictive maintenance, thereby contributing to more sustainable and resilient urban systems.

\section{Conclusion}\label{sec:conclusion}
\noindent
Our study presents a two-stage deep learning framework combining the Temporal Fusion Transformer (TFT) for travel behavior forecasting and a GAN-based model for spatial prediction. The results demonstrate that TFT outperforms traditional time-series models, while the GAN with a 512-dimension latent space yields the most realistic satellite images. The model successfully forecasts long-term mobility trends and generates high-quality spatial representations, making it a valuable tool for transportation planning. Additionally, the model architecture can be adapted for real-time data streams, enabling dynamic decision-making in rapidly changing urban environments, and it will be released for public use. Nonetheless, some limitations remain, including forecasting uncertainties and spatial inconsistencies, which future research can address through enhanced feature engineering and model interpretability improvements.

\section{Acknowledgment}
This research was supported by the Economic Diversifying Research Fund (EDRF) under the grant number (FAR0037938).

\bibliographystyle{IEEEtran}
\bibliography{ref.bib}
\end{document}